\documentclass[runningheads]{llncs}
\usepackage{graphicx}
\usepackage{multirow}
\usepackage{hyperref}
\usepackage[misc]{ifsym}

\usepackage{fancyhdr}
\begin{document}
\title{Dual-FOFE-net Neural Models for Entity Linking with PageRank}
%
%
\author{Feng Wei (\Letter) \and Uyen Trang Nguyen \and Hui Jiang}
\authorrunning{F. Wei et al.}
%
\institute{Department of Electrical Engineering and Computer Science, \\
York University, 4700 Keele St, Toronto, ON M3J 1P3, Canada \\
\email{\{fwei,utn,hj\}@cse.yorku.ca}}
\maketitle              
\thispagestyle{fancy}
\begin{abstract}

This paper presents a simple and computationally efficient approach for entity linking (EL), compared with recurrent neural networks (RNNs) or convolutional neural networks (CNNs), by making use of feedforward neural networks (FFNNs) and the recent dual fixed-size ordinally forgetting encoding (dual-FOFE) method to fully encode the sentence fragment and its left/right contexts into a fixed-size representation. Furthermore, in this work, we propose to incorporate PageRank based distillation in our candidate generation module. Our neural linking models consist of three parts: a PageRank based candidate generation module, a dual-FOFE-net neural ranking model and a simple NIL entity clustering system. Experimental results have shown that our proposed neural linking models achieved higher EL accuracy than state-of-the-art models on the TAC2016 task dataset over the baseline system, without requiring any in-house data or complicated handcrafted features. Moreover, it achieves a competitive accuracy on the TAC2017 task dataset.

\end{abstract}

\section{Introduction}

Named entities (NEs) have received much attention over the last two decades \cite{nadeau2007survey}, mostly focused on recognizing the boundaries of textual NE mentions and classifying them as, e.g., Person (PER), Organization (ORG), Facility (FAC), Geo-political Entity (GPE) or Location (LOC). In 2009, NIST proposed the shared task challenge of EL \cite{mcnamee2009overview}. EL is a similar but broader task than named entity disambiguation (NED). NED is concerned with disambiguating a textual NE mention where the correct entity is known to be one of the Knowledge Base (KB) entries, while EL also requires systems to deal with the case where there is no entry for the NE in the reference KB.

In \cite{ji2011knowledge}, the authors group and summarise different approaches to EL taken by participating systems. There is a vast body of research on NED, highlighted by \cite{ji2016entity}. The problem has been studied extensively by employing a variety of machine learning, and inference methods, including a pipeline of deterministic modules \cite{ling2015design}, simple classifiers \cite{ratinov2011local}, graphical models \cite{durrett2014joint}, classifiers augmented with ILP inference \cite{cheng2013relational}, and more recently, neural approaches \cite{liu2016ustc,sil2017ibm,tang2017zhi,yang2017tai}.

In this paper, we propose to use the recent dual-FOFE method \cite{watcharawittayakul2018dual} to fully encode the left/right contexts for each target mention, and then a simple FFNNs can be trained to make a precise linking for each target mention based on the fixed-size presentation of the contextual information. Moreover, we propose to incorporate PageRank based distillation in our candidate generation system. Compared with \cite{liu2016ustc,tang2017zhi,yang2017tai}, our proposed neural linking models, without requiring any in-house data or complicated handcrafted features, can efficiently achieve higher EL accuracy in terms of computing than the baseline system, and achieves a competitive accuracy on both TAC2016 and TAC2017 task datasets.

The remainder of this paper is organized as follows. Section~\ref{Model} describes our proposed neural linking models. In Section~\ref{Expt}, we discuss experimental results and compare the performance of our proposed models with that of existing state-of-the-art systems. Finally, Section~\ref{Conclusion} draws the conclusions and outlines our future work.

\section{Our Proposed Neural Linking Models}\label{Model}

In this section, we discuss our proposed neural linking models, which consist of three parts: PageRank based candidate generation module, dual-FOFE-net neural ranking model and NIL entity clustering system.

\subsection{PageRank based Candidate Generation}\label{CG}

\begin{figure*}[!ht]
  \centering
  \includegraphics[height=5cm]{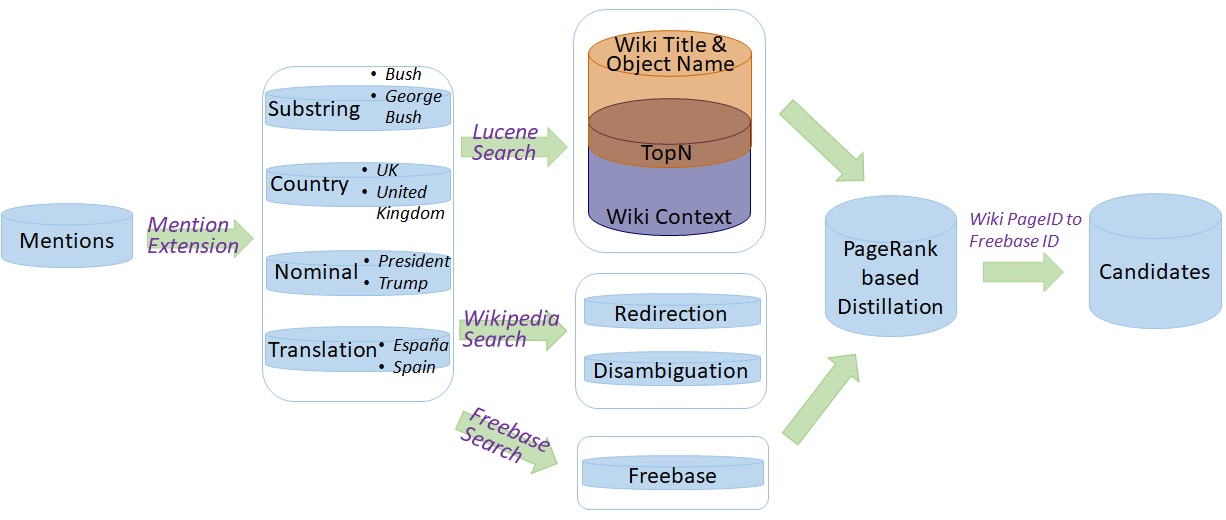}\\
  \caption{The diagram of the PageRank based candidate generation module.}\label{fg:1}
\end{figure*}

Inspired by \cite{page1999pagerank}, we propose to extend the previous work in \cite{liu2016ustc} to incorporate a PageRank based distillation in our candidate generation module to generate candidates for each detected mention. Candidates are generated based on KBs, including \emph{Freebase} and \emph{Wikipedia} \cite{zesch2008extracting}. Lucene fuzzy search strategy is applied in the implementation. The input to this module is a detected mention, and the output is a candidate list, which consists of a group of \emph{Freebase} nodes potentially matching this mention, as shown in Figure \ref{fg:1}.

Following are five types of mention extensions implemented in the candidate generation module:

\begin{itemize}
	\item \emph{Substring Extension}: For each mention, all the recognized named entities in its original context document containing that mention will be selected. For instance, given the mention ``Trump'' in document $d$, ``Donald Trump'' will be selected as its substring extension if the named entity ``Donald Trump'' is found in $d$.
	\item \emph{Translation Extension}: If a mention is in Chinese or Spanish, we invoke Google Translation to obtain its English translation as Translation Extension.
	\item \emph{Country Extension}: The abbreviation of a country name can be extended to a more concrete one. For example, the mention of the geo-political entity ``UK'' will be extended to ``United Kingdom''.
	\item \emph{Nominal Extension}: The nearest recognized entity with the same entity type as its nominal extension will be selected to be added to the query list.
	\item \emph{Traditional Chinese Extension}: If a mention is in the form of simplified Chinese, its traditional Chinese version will be obtained.
\end{itemize}

After the mention extensions, three parallel strategies are applied respectively:
\begin{itemize}
	\item To invoke Lucene fuzzy search on \emph{Wikipedia} titles, first paragraphs, document context and \emph{Freebase} object names.
	\item To query a database with \emph{Wikipedia} redirection and disambiguation information.
	\item To query a database with all \emph{Freebase} entities.
\end{itemize}

\subsubsection{PageRank based Distillation.}

As discussed above, although in general the mention extension step helps to enhance the candidate coverage, it also products too many candidates for a single mention. This behavior leads to much noise, and slows down the whole system.

\begin{figure*}[!ht]
  \centering
  \includegraphics[height=5cm]{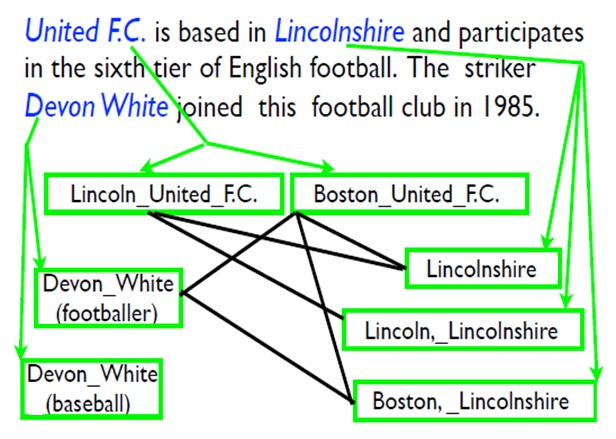}\\
  \caption{A toy document graph for three entity mentions: United F.C., Lincolnshire, Devon White. Source: Adapted from Pershina et al. \cite{pershina2015personalized}}\label{fg:2}
\end{figure*}

In this work, inspired by \cite{page1999pagerank}, we propose to incorporate PageRank based distillation at the last step. As depicted in Figure \ref{fg:2}, a toy document graph includes three entity mentions and seven candidates: three candidates generated for Lincolnshire, and two candidates generated for United F.C. and Devon White each. Each graph node $e(m, c)$ is a pair of an entity mention $m$ and a candidate $c$. An edge is drawn between two candidates of different entities whenever there is a link from the \emph{Wikipedia} page for one candidate to the \emph{Wikipedia} page for the other. There is no edge between candidates competing for the same entity.

It is worth noting that edges in our graph model represent relations between candidates. We insert an edge between two candidates if the \emph{Wikipedia} entry corresponding to either of the two candidates contains a link to the other candidate. We assume that this relation is bidirectional and thus this edge is undirected.

We rank the candidates of each mention based on their outbound link counts to all the recognized mentions in the same document, and keep the top $\tau$ candidates for each mention, where $\tau$ is a distillation factor. We name this step ``distillation''. The score is calculated by the \emph{Wikipedia}'s anchors as follows:

\begin{equation}\label{eq:2}
  score(c) = \sum_{m'}count(c,m')
\end{equation}
where $c$ is the candidate entity, $m'$ is the linked page (a mention identified in the same document) and $count(c,m')$ is the total co-occurrence count of $c$ and $m'$.

Finally, for each detected mention $m$, the candidate generation module generates a list of $K$ candidates as $C=\{c_1,\cdots,c_K\}$, with $K\in[0,\tau]$.

\subsection{Dual-FOFE-net Neural Ranking Models}

\subsubsection{Dual Fixed-Size Ordinally Forgetting Encoding (Dual-FOFE).}

FOFE \cite{zhang2015fixed} was proposed as an alternative to commonly used sequence embedding representations, and achieved competitive results in language modeling. There is a nice theoretical property to guarantee that FOFE codes can almost uniquely encode any variable-length sequence of words into a fixed-size representation without losing any information.

Given a vocabulary $V$, where each word can be represented by a 1-of-$|V|$ one-hot vector. Let $S = \{w_1, \cdots, w_N\}$ denote a sequence of $N$ words from $V$, and $e_n$ denote the one-hot vector of the $n$-th word in $S$, where $1 \leq  n \leq N$. Assuming $z_0 = 0$, the FOFE code $z_n$ of the sequence from word $w_1$ to $w_n$ is shown as follows:
\begin{equation}\label{eq:1}
 z_n = \alpha \cdot z_{n-1} + e_n
\end{equation}
where $\alpha$ is a constant forgetting factor. Thus, $z_n$ can be viewed as a fixed-size representation of the subsequence $\{w_1,\cdots,w_n\} $. We can see that, according to the theoretical properties presented in \cite{zhang2015fixed}, any sequence of variable length can be uniquely and losslessly encoded into a fixed-size representation by FOFE.

This simple ordinally-forgetting mechanism has been applied to some NLP tasks, e.g., \cite{xu2017local,xu2017fofe,xu2016yorknrm} and have achieved very competitive results.

The main idea of dual-FOFE is to generate augmented FOFE encoding codes by concatenating two FOFE codes using two different forgetting factors. Each of these FOFE codes is still computed in the same way as the mathematical formulation shown in Equation (\ref{eq:1}). The difference between them is that we may select to use two different values for the forgetting factor for additional modeling benefits.

\subsubsection{Our Proposed Dual-FOFE-net.}

As described in \ref{CG}, we generate a candidate list $C$ for each detected mention $m$. This list contains a special NIL candidate and some \emph{Freebase} node IDs that match the mention in the candidate generation process. In this work, we propose to use a FFNNs probability ranking model to assign probabilities to all candidates in the list. The candidate with the highest probability is chosen as the final linking result. Each time, the FFNNs probability ranking model takes a mention $m$ and a candidate $c_k$ from the list $C$ to compute a matching score, $e_k$. In order to do this, we make use of dual-FOFE to encode mention context features for the neural network.

\begin{figure*}[ht!]
  \centering
  \includegraphics[height=6.5cm]{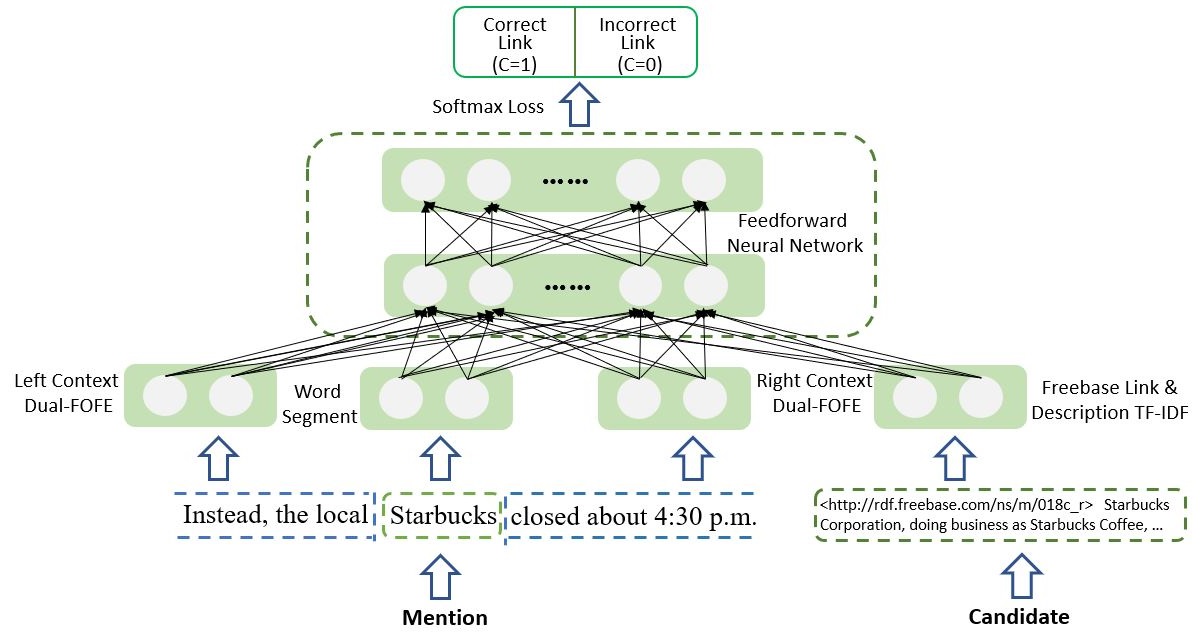}\\
  \caption{Our proposed neural ranking model using dual-FOFE codes as input and a feed-forward neural network.}\label{fg:3}
\end{figure*}

As shown in Figure \ref{fg:3}, the input feature vector to the FFNNs probability ranking model is a concatenation of all the following features:

\begin{itemize}
  \item \emph{Mention string embedding}: Each detected mention is represented as a bag-of-words vector. This bag-of-words vector is projected into a 128-dimension dense vector.
  \item \emph{Document context}: The left and right contexts of each mention are encoded by dual-FOFE, and projected into a 256-dimension dense vector.
  \item \emph{Knowledge base description}: The corresponding KB, \emph{Freebase}, description of each candidate and target mention is individually represented as one bag-of-words vectors (weighted using the TFIDF schema), which is mapped to a 128-dimension dense vector. As for Chinese and Spanish, since the languages have fewer resources than English in \emph{Freebase}, we invoke \emph{Google} APIs, which extract the translation to expand their Chinese and Spanish descriptions separately.
\end{itemize}

In this paper, we use the rectified linear activation function, i.e., $f(x)=\max(0,x)$, to compute from activations to outputs in each hidden layer, which are in turn fed to the next layer as inputs. For the output layer, we make use of the softmax function to compute posterior probabilities between two nodes, standing for correct links or incorrect links, shown as follows:
\begin{equation}\label{eq:3}
  P_r(c_k|m) = \frac{exp(e_k)}{\sum_{k=1}^{K}exp(e_k)}.
\end{equation}

\subsection{NIL Entity Clustering}

For all mentions identified as NIL by the above dual-FOFE-net neural ranking model, we perform a simple rule-based algorithm to cluster them: Different named NIL mentions are grouped into one cluster only if their mention strings are the same (case-insensitive).

\section{Experiments and Results}\label{Expt}

In this section, we evaluate the effectiveness of our proposed methods on the benchmark datasets: TAC2016 and TAC2017 Trilingual Entity Discovery and Linking (EDL) tasks, and compare the performance of our proposed neural linking model with state-of-the-art models \cite{liu2016ustc,yang2017tai} on both TAC2016 and TAC2017 task datasets.

In \cite{liu2016ustc}, the authors use some in-house data annotated by themselves, which consists of about 10,000 Chinese and English documents acquired through their web crawler. These documents are internally labelled using some annotation rules similar to the KBP guidelines. In \cite{yang2017tai}, many complicated handcrafted features are created, including the mention level feature, entity level feature, mention-to-entity feature and entity-to-entity feature. It is worth noting that we have not used any in-house data or handcrafted features in our models, and all used features (either word or character level) are automatically derived from the data based on the simple FOFE formula.

\subsection{Dataset}

\begin{table}[t!]
  \caption{\label{tb:1}Number of Documents in TAC2015-2017}
  \centering
  \begin{tabular}{|c|c|c|c|c|}
    \hline      
                          & English & Chinese & Spanish & ALL \\
    \hline      
                           15 Train & 168   & 147     & 129    & 444   \\
    \hline      
                           15 Eval   & 167  & 166     & 167    & 500   \\
    \hline      
                           16 Eval   & 168  & 167     & 168    & 503   \\
   \hline      
                           17 Eval   & 167  & 167     & 166    & 500   \\
    \hline      
  \end{tabular}
\end{table}

Given a document collection in three languages (English, Chinese and Spanish), the TAC trilingual EDL task \cite{ji2016overview,ji2017overview} automatically identifies entities from a source collection of textual documents in multiple languages, as shown in Table \ref{tb:1}, classifies them into one of the following pre-defined five types: Person (PER), Geo-political Entity (GPE), Organization (ORG), Location (LOC) and Facility (FAC), links them to an existing KB (BaseKB)\footnote{http://basekb.com/}, and clusters mentions for those NIL entities that do not have corresponding BaseKB entries. The corpus consists of news articles and discussion forum posts published in recent years, related to but non-parallel across languages.

\subsection{Neural Model Setup}
Three models are trained and evaluated independently. Three sets of word embeddings of 128 dimensions are derived from English Gigaword \cite{parker2011english}, Chinese Gigaword \cite{graff2005chinese} and Spanish Gigaword \cite{mendonca2009spanish}, respectively. Since Chinese segmentation is not reliable, based on a predefined set of all possible characters, we view the focus token as a character sequence and encode it using dual-FOFE. We then project the character encodings using a trainable character embedding matrix.

Relying on the development experiments, the set of hyper-parameters used in our experiments is summarized as follows: i) Learning rate: All models are trained using the stochastic gradient descent (SGD) algorithm while the learning rate is set to be 0.1; ii) Network structure: Three hidden layers and ReLUs \cite{nair2010rectified} as the nonlinear activation function,  randomly initialized based on a uniform distribution between $-\sqrt{\frac{6}{N_i+N_o}}$ and $\sqrt{\frac{6}{N_i+N_o}}$ \cite{glorot2011deep}; iii)  Dropout \cite{hinton2012improving} is adopted during training; iv) Number of epochs: 30; v) Chinese character embeddings: 64 dimensions, randomly initialized; vi) Forgetting factor: $\alpha$ = (0.5, 0.9) \footnote{The choice of the forgetting factors $\alpha$ is empirical. We've evaluated on a development set in some early experiments. It turns out that $\alpha$ = (0.5, 0.9) is the best. As a result, (0.5, 0.9) is used for all EL tasks throughout this paper.}; vii) Distillation factor: $\tau = 20$.

\subsection{Evaluation Metrics}

To evaluate the effectiveness of our proposed models, we use the standard NERLC and CEAFmC metrics, which are combined measures of linking and clustering performance. \footnote{More details regarding data format and scoring metric can be found in http://nlp.cs.rpi.edu/kbp/}

\subsection{Results and Discussion}
Table \ref{tb:2} shows the performance of our proposed model on the TAC2016 task dataset along with the TAC Rank 1 system \cite{liu2016ustc}. Our model outperforms the other system by 1.2\% in terms of NERLC, and by 1.1\% in terms of CEAFmC with the overall trilingual EL performance. Furthermore, each of the three individual models is better than its counterpart in the TAC Rank 1 system in terms of both NERLC and CEAFmC. As for the TAC2017 task dataset, shown in Table \ref{tb:3}, encouragingly, the NERLC performance of English, Chinese and trilingual overall, outperforms the best system \cite{yang2017tai}, by 0.4\%, 0.6\% and 0.2\% separately, and the CEAFmC performance of English, Spanish and trilingual overall, is slightly better than the best system \cite{yang2017tai}, by 0.5\%, 0.4\% and 0.4\% separately.

Following are the advantages of our proposed models over the state-of-the-art on both TAC2016 and TAC2017 task datasets. First, unlike the systems in \cite{liu2016ustc,yang2017tai}, our models do not rely on any in-house data or complicated handcrafted features. It is very time consuming and labour intensive to prepare clean annotated in-house data, or to collect and select good handcrafted features. More importantly, we have not used any handcrafted features in our models, and all used features (either word or character level) are automatically derived from the data based on the simple FOFE formula. Secondly, we present a simple and computationally efficient approach compared with recurrent neural networks (RNNs) or convolutional neural networks (CNNs), by making use of feedforward neural networks (FFNNs) and the recent dual fixed-size ordinally forgetting encoding (dual-FOFE) method to fully encode the sentence fragment and its left/right contexts into a fixed-size representation. Feedforward neural networks (FFNNs) use rather simple structures consisting of several fully-connected layers. These neural networks are known to be powerful as universal approximators~\cite{hornik1991approximation}, and they are simpler and faster to train and inference than the more recent variants such as long short-term memory (LSTM)~\cite{hochreiter1997long}, Gated Recurrent Unit (GRU)~\cite{cho2014learning} or CNNs. Thus, our proposed dual-FOFE-net neural model is light and highly efficient compared with RNNs or CNNs. Last but not least, our proposed PageRank based distillation not only enhances the candidate coverage, but also speeds up the whole models.

\begin{table}[t!]
  \caption{\label{tb:2}Performance comparison with the best system on the TAC2016 datasets (in terms of NERLC $F_1$ and CEAFmC $F_1$).}
  \centering
  \begin{tabular}{|c|c|c|c|c|}
    \hline     
                             &    \multicolumn{2}{c|}{\begin{tabular}{c}\cite{liu2016ustc} \\TAC Rank 1\end{tabular}}   &  \multicolumn{2}{c|}{\begin{tabular}{c}Our proposed \\models\end{tabular}}  \\
    \hline   \hline   
                             &  NERLC      &    CEAFmC  &       NERLC              &    CEAFmC      \\
    \hline
          Trilingual         &  64.7       &      66.0  &       \textbf{65.9}      &  \textbf{67.1} \\
    \hline      
          English            &  66.6       &      67.6  &       \textbf{67.7}      &  \textbf{69.0} \\
    \hline      
          Chinese            &  65.0       &      70.2  &       \textbf{66.4}      &  \textbf{70.7} \\
    \hline      
          Spanish            &  61.6       &      63.5  &       \textbf{62.5}      &  \textbf{64.4} \\
    \hline      
  \end{tabular}
\end{table}

\begin{table}[t!]
  \caption{\label{tb:3}Performance comparison with the best system on the TAC2017 datasets (in terms of NERLC $F_1$ and CEAFmC $F_1$).}
  \centering
  \begin{tabular}{|c|c|c|c|c|}
    \hline     
                             &    \multicolumn{2}{c|}{\begin{tabular}{c}\cite{yang2017tai} \\TAC Rank 1\end{tabular}}   &  \multicolumn{2}{c|}{\begin{tabular}{c}Our proposed \\models\end{tabular}}  \\
    \hline   \hline   
                             &  NERLC      &    CEAFmC  &       NERLC              &    CEAFmC      \\
    \hline
          Trilingual         &  67.8       &      70.5  &       \textbf{68.0}      &  \textbf{70.9} \\
    \hline      
          English            &  66.8       &      68.8  &       \textbf{67.2}      &  \textbf{69.3} \\
    \hline      
          Chinese            &  71.0       &      73.2  &       \textbf{71.6}      &  72.4          \\
    \hline      
          Spanish            &  65.0       &      68.9  &       64.8               &  \textbf{69.3} \\
    \hline      
  \end{tabular}
\end{table}

\section{Conclusion}\label{Conclusion}

This paper presents a simple and computationally efficient approach to EL by applying FFNNs on top of dual-FOFE features. Furthermore, we propose to incorporate PageRank based distillation in our candidate generation module. Our experiments have shown that, without requiring any in-house data or complicated handcrafted features, it achieves higher EL accuracy than state-of-the-art systems on the TAC2016 task dataset, and offers a competitive accuracy on the TAC2017 task dataset.

In our future work, we will evaluate our neural linking models on more datasets and conduct more experiments to measure the sensitivity of the system to the values of some hyperparameters (e.g., number of hidden layers). In addition, we will explore more architectures (e.g., convolutional layers), to quantify the contribution of some modules.

%
%
%
%
\bibliography{icann2019}
\bibliographystyle{splncs04}
\end{document}